\documentclass[a4paper,twoside]{article}

\usepackage{epsfig}
\usepackage{subcaption}
\usepackage{calc}
\usepackage{amssymb}
\usepackage{amstext}
\usepackage{amsmath}
\usepackage{amsthm}

\usepackage{multicol}
\usepackage{pslatex}
\usepackage{apalike}
\usepackage{algorithm2e}
\usepackage[bottom]{footmisc}
\DeclareMathOperator{\med}{med}
\usepackage{graphicx}
\usepackage{tikz}
\usetikzlibrary{
  arrows.meta,
  calc,
  positioning,
  fit,
  backgrounds
}
\usepackage{booktabs}
\usepackage{multirow}

\usepackage{SCITEPRESS}     

\begin{document}

\title{DynEoMT: Learning Object Dynamicity from Online Segmentation Queries}

\author{\authorname{Calvin Galagain\sup{1,2}\orcidAuthor{0009-0004-2168-9617}, Martyna Poreba\sup{1}\orcidAuthor{0000-0002-5102-7735} and François Goulette\sup{2}\orcidAuthor{0000-0003-1527-2650}}
\affiliation{\sup{1}Université Paris-Saclay, CEA, List, F-91120, Palaiseau, France}
\affiliation{\sup{2}U2IS, ENSTA Paris, Institut Polytechnique de Paris, 91120, Palaiseau, France
}
\email{\{calvin.galagain, martyna.poreba\}@cea.fr, \{francois.goulette\}@ensta.fr}
}


\abstract{Video segmentation models delineate and associate regions over time, but do not explicitly indicate whether they move independently of the camera. Query-based architectures propagate region representations across frames, raising a central question: can existing segmentation queries also support dynamicity prediction? We introduce DynEoMT, which adds a lightweight dynamicity head to existing segmentation queries without altering the segmentation or query-propagation architecture. To supervise this additional task, we generate class-agnostic pseudo-labels offline from camera-compensated optical flow. At inference, DynEoMT requires only the current frame and propagated queries, without explicit motion or geometric inputs. On VIPSeg, an MLP probe trained on frozen ViT-L queries achieves conditional balanced accuracy within 0.05 percentage points of full fine-tuning, suggesting that dynamicity-relevant information is already accessible in these representations. Across VIPSeg, OVIS, YouTube-VIS 2022, and VSPW, the ViT-B variant achieves conditional balanced accuracies of 84.30\%, 67.98\%, 68.59\%, and 87.64\%, respectively, while largely preserving segmentation performance with minimal overhead in the measured inference path.}

\onecolumn \maketitle \normalsize \setcounter{footnote}{0} \vfill

\section{\uppercase{Introduction}}
\label{sec:introduction}

Reliable perception in dynamic scenes requires distinguishing independently moving objects from apparent image motion induced by camera ego-motion. Semantics alone are insufficient: pedestrians may remain stationary and vehicles may be parked. Conversely, static structures can exhibit substantial image displacement as the camera moves. In robotic mapping and planning, confusing object motion with camera-induced motion can degrade localization and scene representation ~\cite{bescos2018dynaslam,GALAGAIN2026105715}.

\begin{figure}[ht]
  \centering
  \resizebox{\columnwidth}{!}{
  \begin{tikzpicture}[x=1cm,y=1cm,font=\sffamily,text=black!82]
    \definecolor{archInk}{RGB}{31,37,48}
    \definecolor{archMuted}{RGB}{91,101,116}
    \definecolor{archBlue}{RGB}{43,123,238}
    \definecolor{archBlueFill}{RGB}{236,243,255}
    \definecolor{archGreen}{RGB}{28,157,84}
    \definecolor{archGreenFill}{RGB}{233,248,239}
    \definecolor{archOrange}{RGB}{242,101,34}
    \definecolor{archOrangeFill}{RGB}{255,242,232}
    \definecolor{archRed}{RGB}{211,48,64}
    \definecolor{archRedFill}{RGB}{255,238,241}
    \tikzset{
      archArrow/.style={-{Latex[length=2.0mm,width=1.35mm]},draw=archInk,line width=0.75pt},
      archLine/.style={draw=archInk,line width=0.75pt},
      archGreenArrow/.style={-{Latex[length=2.0mm,width=1.35mm]},draw=archGreen,line width=0.8pt},
      archRedArrow/.style={-{Latex[length=2.0mm,width=1.35mm]},draw=archRed,line width=0.85pt},
      archBox/.style={rounded corners=1.5pt,line width=0.75pt,align=center,font=\small,text=archInk}
    }

    \node[draw=archInk,line width=0.6pt,inner sep=0pt] (image) at (1.05,2.62)
      {\includegraphics[width=1.95cm,height=1.09cm]{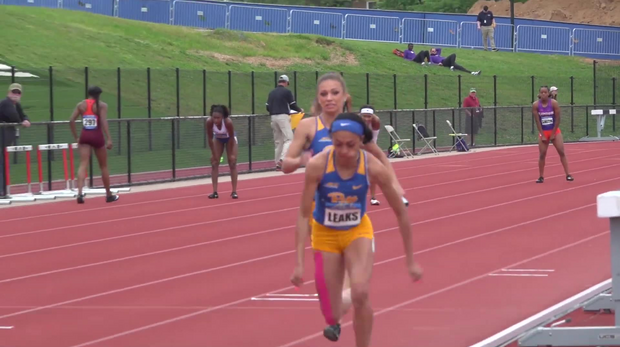}};
    \node[font=\scriptsize,text=archMuted] at (1.05,3.34) {current frame $I_t$};
    \node[archBox,draw=archOrange,fill=archOrangeFill,minimum width=1.95cm,minimum height=0.72cm]
      (vit) at (1.05,1.50) {\textbf{ViT encoder}};
    \node[archBox,draw=archGreen,fill=archGreenFill,minimum width=1.85cm,minimum height=0.56cm]
      (prevq) at (1.05,0.45) {propagated\\[-1pt]queries $Q_{t-1}$};
    \draw[archArrow] (image.south) -- (vit.north);
    \draw[archGreenArrow] (prevq.north) -- (vit.south);

    \node[archBox,draw=archGreen,fill=archGreenFill,minimum width=1.45cm,minimum height=0.58cm]
      (qt) at (3.55,1.50) {queries\\[-1pt]$Q_t$};
    \draw[archGreenArrow] (vit.east) -- (qt.west);
    \node[archBox,draw=archGreen,fill=archGreenFill,minimum width=1.45cm,minimum height=0.42cm,
      font=\scriptsize] (nextq) at (3.55,0.55) {propagate};
    \draw[archGreenArrow] (qt.south) -- (nextq.north);

    \node[archBox,draw=archBlue,fill=archBlueFill,minimum width=2.28cm,minimum height=0.72cm]
      (seg) at (6.10,2.20) {class + mask\\[-1pt]heads};
    \node[archBox,draw=archRed,fill=archRedFill,minimum width=2.28cm,minimum height=0.72cm]
      (dyn) at (6.10,0.92) {dynamicity\\[-1pt]head};
    \coordinate (qtseg) at ([yshift=0.11cm]qt.east);
    \coordinate (qtdyn) at ([yshift=-0.11cm]qt.east);
    \draw[archArrow] (qtseg) -- (seg.west);
    \draw[archRedArrow] (qtdyn) -- (dyn.west);

    \node[draw=archBlue,line width=0.65pt,inner sep=0pt] (segout) at (8.85,2.42)
      {\includegraphics[width=2.20cm,height=1.23cm]{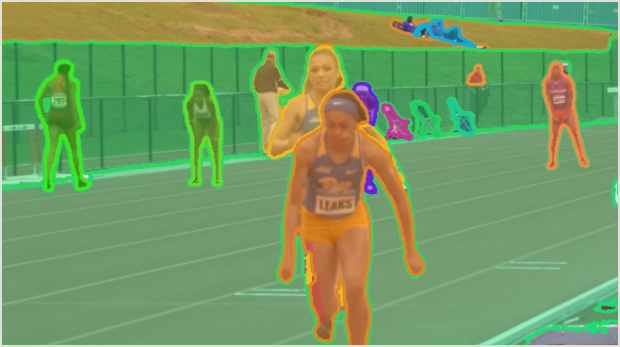}};
    \draw[archArrow] (seg.east) -- (segout.west);
    \node[font=\scriptsize,text=archMuted] at (8.85,1.65) {classes, masks, IDs};
    \node[draw=archRed,line width=0.65pt,inner sep=0pt] (dynout) at (8.85,0.83)
      {\includegraphics[width=2.20cm,height=1.23cm]{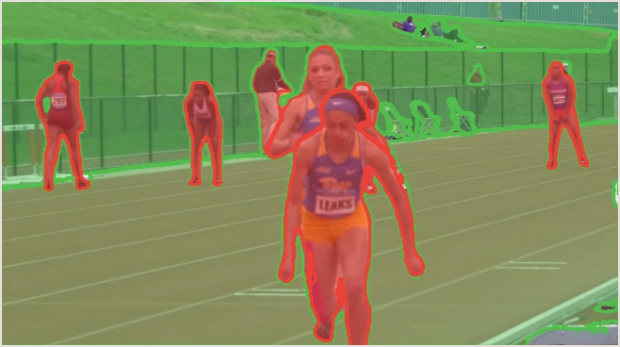}};
    \draw[archRedArrow] (dyn.east) -- (dynout.west);
    \node[font=\scriptsize,text=archMuted] at (8.85,0.06) {dynamic / static};
  \end{tikzpicture}}
  \caption{Overview of DynEoMT. The current image and the queries propagated
  from the previous frame are jointly processed by a ViT. The resulting current
  queries feed the original segmentation heads and the added dynamicity head,
  producing segmentation outputs and dynamic/static predictions.}
  \label{fig:teaser}
\end{figure}

Video segmentation provides spatial and temporal scene representations through semantic, instance, and panoptic segmentation. Recent query-based architectures represent segmented regions using learned embeddings. Among them, VidEoMT~\cite{norouzi2026videomt} propagates segmentation queries across frames and jointly processes them with current image patches in an encoder-only Vision Transformer (ViT). These queries support segmentation and temporal association, but do not explicitly predict whether the represented regions move independently of the camera. This raises a central question: can existing segmentation queries also support dynamicity prediction without explicit motion estimation at inference?

We investigate this question with DynEoMT, which extends VidEoMT with a lightweight dynamicity head operating directly on segmentation queries. The original segmentation and query-propagation paths remain unchanged. Since existing video segmentation datasets lack dynamicity annotations, we generate class-agnostic pseudo-labels from camera-compensated optical flow to supervise the additional prediction task. At inference, DynEoMT requires only the current frame and propagated queries, without optical flow, depth, camera pose, or retained previous-frame features. Unlike dedicated moving-object segmentation (MOS) methods ~\cite{xie2024flowsam,huang2025segmentanymotion,goli2025romo,xie2026gmos}, our approach assigns dynamicity directly to regions produced by standard video segmentation tasks.

Our contributions are threefold:
\begin{itemize}
    \item DynEoMT, a lightweight extension of query-based video segmentation that predicts region-level dynamicity directly from segmentation queries, without explicit motion inputs at inference;

    \item A class-agnostic pseudo-label generation approach based on camera-compensated optical flow, enabling dynamicity supervision across semantic, instance, and panoptic video segmentation;

    \item An evaluation across four video segmentation benchmarks, including the proposed DynSeg-F1 metric and probing experiments that examine the dynamicity information available in segmentation queries.
\end{itemize}

\section{\uppercase{Related Work}}
\label{sec:sota}

\paragraph{Query-based video segmentation.} 
Modern query-based video segmentation methods combine mask prediction with temporal object association. MinVIS~\cite{huang2022minvis} associates independently predicted image queries across frames, whereas DVIS and its successors decouple frame-level segmentation, tracking, and temporal refinement~\cite{zhang2023dvis,zhang2025dvispp,zhou2024dvisdaq}. Online methods instead maintain object-level states across frames, as in InsPro~\cite{he2022inspro}, CAROQ~\cite{choudhuri2023caroq}, and CAVIS~\cite{lee2025cavis}. EoMT~\cite{kerssies2025eomt} introduces an encoder-only image segmentation architecture in which a pretrained plain ViT jointly processes image patches and segmentation queries. VidEoMT~\cite{norouzi2026videomt} extends this formulation to video by jointly processing propagated object queries and current image patches, supporting segmentation and temporal association without a dedicated tracker.

\paragraph{Moving-object segmentation.}
Moving-object segmentation (MOS) aims to discover and delineate objects whose motion is independent of the observing camera. Earlier approaches use optical flow for motion-based grouping~\cite{yang2021motiongrouping}. Recent methods combine motion cues with pretrained visual representations. FlowSAM~\cite{xie2024flowsam} integrates optical flow with SAM~\cite{kirillov2023segmentanything}, while SegAnyMo~\cite{huang2025segmentanymotion} leverages long-range point trajectories, DINO features, and SAM2~\cite{ravi2025sam2}. RoMo~\cite{goli2025romo} fuses optical flow and epipolar constraints, whereas GeoMotion~\cite{he2026geomotion} predicts motion masks from latent 4D geometry. Most recently, GMOS~\cite{xie2026gmos} combines geometric and SAM2 representations in a proposer--propagator framework to predict temporally varying object motion states, and introduces GMOS-2K and the MOS-I evaluation protocol. However, these methods produce motion-centric regions rather than assigning a motion state to regions defined by standard semantic, instance, or panoptic video segmentation.

\paragraph{Motion cues and geometric reasoning.}
Optical flow provides dense image-plane correspondences, with RAFT~\cite{teed2020raft} operating on frame pairs and MemFlow~\cite{dong2024memflow} incorporating temporal information. However, observed flow entangles camera motion, scene geometry, independently moving objects, occlusions, and estimation errors. Geometry-aware approaches therefore combine motion cues with depth, camera pose, and rigidity constraints. RigidMask~\cite{yang2021rigidmask} jointly estimates depth, camera motion, and rigid-motion segments, whereas EffiScene~\cite{jiao2021effiscene} predicts optical flow, depth, camera pose, and per-pixel rigidity. Long-term point trajectories have also been used as supervision for learning object segmentation~\cite{karazija2024learning}.

\section{\uppercase{DynEoMT}}

\begin{figure*}[t]
  \centering
  \resizebox{\textwidth}{!}{
  \begin{tikzpicture}[x=1cm,y=1cm,font=\sffamily,text=black!80]
  \definecolor{archInk}{RGB}{31,37,48}
  \definecolor{archMuted}{RGB}{91,101,116}
  \definecolor{archOrange}{RGB}{242,101,34}
  \definecolor{archOrangeFill}{RGB}{255,242,232}
  \definecolor{archBlue}{RGB}{43,123,238}
  \definecolor{archBlueFill}{RGB}{236,243,255}
  \definecolor{archGreen}{RGB}{28,157,84}
  \definecolor{archGreenFill}{RGB}{233,248,239}
  \definecolor{archRed}{RGB}{211,48,64}
  \definecolor{archRedFill}{RGB}{255,238,241}
  \tikzset{
    archArrow/.style={-{Latex[length=2.1mm,width=1.4mm]},draw=archInk,line width=0.8pt,shorten <=1pt,shorten >=1pt},
    archSoftArrow/.style={-{Latex[length=2mm,width=1.3mm]},draw=archMuted,line width=0.7pt,shorten <=1pt,shorten >=1pt},
    archGreenArrow/.style={-{Latex[length=2.1mm,width=1.4mm]},draw=archGreen,line width=0.85pt,shorten <=1pt,shorten >=1pt},
    archBus/.style={draw=archMuted,line width=0.75pt},
    archVitBlock/.style={draw=archOrange,fill=archOrangeFill,rounded corners=1.5pt,line width=0.8pt,
      minimum width=3.75cm,minimum height=0.65cm,align=center,font=\small,text=archInk},
    archBlueBox/.style={draw=archBlue,fill=archBlueFill,rounded corners=1.5pt,line width=0.8pt,
      minimum width=1.82cm,minimum height=0.65cm,align=center,font=\small,text=archInk},
    archGreenBox/.style={draw=archGreen,fill=archGreenFill,rounded corners=1.5pt,line width=0.8pt,
      minimum width=1.85cm,minimum height=0.72cm,inner xsep=4.5pt,inner ysep=3pt,
      align=center,font=\small,text=archInk},
    archRedBox/.style={draw=archRed,fill=archRedFill,rounded corners=1.5pt,line width=0.9pt,
      minimum width=1.90cm,minimum height=0.65cm,align=center,font=\small,text=archInk},
    archPatchToken/.style={draw=archBlue,fill=archBlueFill,rounded corners=0.45pt,
      minimum size=0.20cm,inner sep=0pt,line width=0.48pt},
    archQueryToken/.style={draw=archGreen,fill=archGreenFill,rounded corners=0.45pt,
      minimum size=0.20cm,inner sep=0pt,line width=0.48pt},
    archTokenStrip/.style={draw=archInk!50,fill=white,rounded corners=1pt,line width=0.55pt,
      minimum height=0.38cm,inner sep=0pt},
    archImage/.style={draw=archInk,line width=0.65pt,inner sep=0pt},
    archLabel/.style={font=\scriptsize,text=archMuted,align=center}
  }
  \node[archImage] (current) at (1.05,3.55)
    {\includegraphics[width=2.05cm,height=1.15cm]{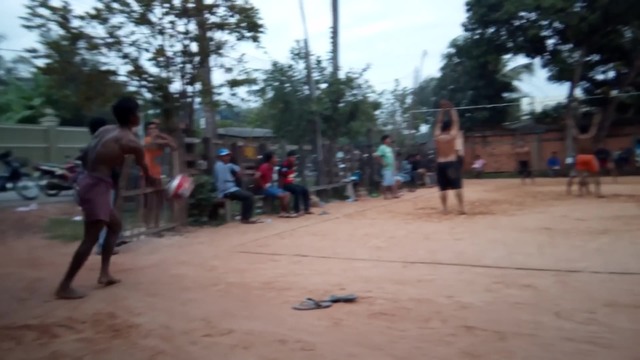}};
  \node[archLabel,below=1.6mm of current] {current image $I_t$};
  \node[archBlueBox,minimum width=1.45cm] (embed) at (3.18,3.55) {Patch\\embedding};
  \draw[archArrow] (current.east) -- (embed.west);
  \node[archGreenBox,text width=1.86cm,minimum width=2.18cm,minimum height=0.74cm]
    (previous) at (1.15,1.32) {Track queries\\[-1pt]$Q_{t-1}$};
  \node[archTokenStrip,minimum width=1.60cm] (prevstrip) at (1.15,0.68) {};
  \foreach \i in {0,...,4} {
    \node[archQueryToken] (prevtoken\i) at (0.55+0.30*\i,0.68) {};
  }
  \node[archGreenBox,text width=1.24cm,minimum width=1.62cm,minimum height=0.78cm]
    (fusion) at (3.46,1.32) {Query\\fusion};
  \draw[-{Latex[length=1.7mm,width=1.15mm]},draw=archGreen,line width=0.85pt,shorten <=1pt,shorten >=1pt]
    (previous.east) -- (fusion.west);
  \node[font=\scriptsize\bfseries,text=archBlue] (contextlabel) at (2.03,0.25)
    {only temporal context};
  \begin{scope}[on background layer]
    \node[draw=archBlue,fill=archBlueFill,rounded corners=2pt,line width=0.7pt,
      fit=(previous)(fusion)(prevstrip)(contextlabel),inner sep=2.5pt] {};
  \end{scope}
  \node[draw=archOrange,fill=archOrangeFill,rounded corners=3pt,line width=1pt,
    minimum width=4.05cm,minimum height=4.35cm] (vit) at (6.65,2.45) {};
  \node[font=\normalsize\bfseries,anchor=north] at ([yshift=-2mm]vit.north)
    {Pretrained ViT encoder};
  \node[archVitBlock] (lone) at (6.65,3.55) {Encoder blocks $\times L_1$};
  \node[archVitBlock] (ltwo) at (6.65,1.50) {Joint encoder blocks $\times L_2$};
  \node[archTokenStrip,minimum width=3.66cm,minimum height=0.38cm] (vittokenstrip) at (6.71,2.60) {};
  \draw[draw=archInk!35,line width=0.55pt] (7.16,2.43) -- (7.16,2.77);
  \foreach \i in {0,...,6} {
    \node[archPatchToken] at (5.10+0.30*\i,2.60) {};
  }
  \foreach \i in {0,...,3} {
    \node[archQueryToken] at (7.42+0.30*\i,2.60) {};
  }
  \draw[archArrow] (embed.east) -- (lone.west);
  \draw[archSoftArrow] (lone.south) -- (6.65,2.79);
  \draw[archGreenArrow] (fusion.east) -- (vittokenstrip.west);
  \draw[archSoftArrow] (6.65,2.41) -- (ltwo.north);
  \node[archGreenBox,text width=1.58cm,minimum width=1.88cm,minimum height=0.72cm] (query) at (10.10,2.70)
    {Queries $Q_t$};
  \node[archTokenStrip,minimum width=1.22cm] (querystrip) at (10.10,2.07) {};
  \foreach \i in {0,...,3} {
    \node[archQueryToken] at (9.65+0.30*\i,2.07) {};
  }
  \draw[archGreenArrow] (ltwo.east) -- (query.west);
  \node[archGreenBox,minimum width=1.85cm,minimum height=0.55cm,font=\scriptsize]
    (propagate) at (10.10,0.95) {propagate to $t+1$};
  \draw[archGreenArrow] (querystrip.south) -- (propagate.north);
  \node[archBlueBox] (classhead) at (12.55,2.70) {Class head};
  \node[archBlueBox] (maskhead) at (14.95,2.70) {Mask head};
  \node[archRedBox] (dynhead) at (17.35,2.70) {Dynamicity head};
  \node[archBlueBox] (classout) at (12.55,1.35) {Class logits\\[-1pt]$p^c_{i,t}$};
  \node[archBlueBox] (maskout) at (14.95,1.35) {Mask output\\[-1pt]$\hat M_{i,t}$};
  \node[archRedBox] (dynout) at (17.35,1.35) {Dynamicity score\\[-1pt]$\hat p^d_{i,t}$};
  \coordinate (busleft) at (11.25,2.70);
  \coordinate (busclass) at (12.55,3.49);
  \coordinate (busmask) at (14.95,3.49);
  \coordinate (busdyn) at (17.35,3.49);
  \node[draw=archRed,dashed,rounded corners=2pt,line width=0.8pt,
    fit=(dynhead)(dynout),inner sep=4pt] {};
  \draw[archBus] (query.east) -- (busleft) -- (11.25,3.49) -- (busdyn);
  \draw[archSoftArrow] (busclass) -- (classhead.north);
  \draw[archSoftArrow] (busmask) -- (maskhead.north);
  \draw[-{Latex[length=2mm,width=1.3mm]},draw=archRed,line width=0.75pt,shorten <=1pt,shorten >=0pt]
    (busdyn) -- (dynhead.north);
  \draw[archSoftArrow] (classhead) -- (classout);
  \draw[archSoftArrow] (maskhead) -- (maskout);
  \draw[-{Latex[length=2mm,width=1.3mm]},draw=archRed,line width=0.75pt,shorten <=0pt,shorten >=0pt]
    (dynhead.south) -- (dynout.north);

  \end{tikzpicture}}
   \caption{DynEoMT architecture. Current image patches and object queries propagated from the preceding frame are jointly processed by the ViT encoder. The resulting queries feed the original classification and mask heads together with the proposed query-level dynamicity head, and are then propagated to the next frame. The propagated queries constitute the only temporal context available at inference. The dynamicity branch is the sole architectural addition to VidEoMT; no previous or future RGB frames, optical flow, depth, or camera pose are required.}
  \label{fig:architecture}
  \end{figure*}

DynEoMT extends VidEoMT~\cite{norouzi2026videomt} with region-level dynamicity prediction, as illustrated in Figure~\ref{fig:architecture}. A lightweight prediction head operates directly on the existing segmentation queries, while the original segmentation and query-propagation paths remain unchanged.

\subsection{Online Propagated-Query Architecture}

VidEoMT represents segmentation regions using $N$ query embeddings of dimension $D$. At the first frame, the learned queries $Q_0\in\mathbb{R}^{N\times D}$ are jointly processed with the image patch tokens. At each subsequent frame, the queries from the preceding step are updated and combined with the learned initialization

\begin{gather}
\widetilde Q_{t-1}
=
W_u Q_{t-1}+Q_0,
\label{eq:query_update}
\\
X_t
=
[\widetilde Q_{t-1};P(I_t)],
\label{eq:input_tokens}
\\
Q_t,P'_t
=
\operatorname{ViT}(X_t).
\label{eq:vit}
\end{gather}
where $P(I_t)$ denotes the patch embeddings of the current frame and $W_u$ is the query-update projection. The resulting queries $Q_t$ feed the classification and mask heads and are propagated to the next frame. This processing path is unchanged from VidEoMT.

At inference, each step processes only the current frame $I_t$ and the propagated queries $Q_{t-1}$; previous image tokens and feature maps are not retained. Although multiple frames are unrolled during training, each step accesses only the current image and the queries emitted by the preceding step.

\subsection{Query-Level Dynamicity Prediction}

We predict the dynamicity of each query-represented
region using a lightweight two-layer MLP. For each
current query $q_{i,t}\in\mathbb{R}^{D}$, the head computes

\begin{align}
\ell^d_{i,t}
&= W_2\operatorname{GELU}(W_1q_{i,t}+b_1)+b_2,
\label{eq:head_logit}
\\
\hat p^d_{i,t}
&= \operatorname{sigmoid}(\ell^d_{i,t}).
\label{eq:head_prob}
\end{align}

The hidden layer has width 256. The dynamicity head operates in parallel with the original classification and mask heads, without using predicted masks or explicit motion or geometric inputs. The same head is applied to auxiliary segmenter outputs for deep supervision.

\subsection{Joint Assignment and Loss}

DynEoMT preserves the original Hungarian assignment~\cite{carion2020detr}, computed from class, point-sampled mask BCE, and Dice costs. Dynamicity therefore cannot influence which query is assigned to an annotated mask. For each annotated region, we use a dynamicity target $y^d_{j,t}\in\{1,0,\varnothing\}$, denoting dynamic, static, or uncertain states, respectively. These targets are generated offline as described in Section~\ref{sec:offline_labels}. Let $\mathcal{M}_v$ denote the matched query--target pairs with valid dynamicity labels. To address the dynamic/static imbalance, we optimize the weighted binary cross-entropy

\begin{equation}
\begin{aligned}
\mathcal L_d
={}&-\frac{1}{|\mathcal M_v|}
\sum_{(i,j)\in\mathcal M_v}
\Big[
w_+ y^d_{j,t}\log\hat p^d_{i,t}
\\
&\qquad
+(1-y^d_{j,t})\log(1-\hat p^d_{i,t})
\Big].
\end{aligned}
\label{eq:dynamicloss}
\end{equation}

Uncertain targets are excluded from $\mathcal{M}_v$. When no valid matches are available, the dynamicity loss is set to zero. The complete objective is

\begin{equation}
 \mathcal L=2\mathcal L_{cls}+5\mathcal L_{mask}+5\mathcal L_{dice}
 +\mathcal L_d,
\end{equation}

applied to the final prediction and all auxiliary segmenter outputs. Dynamicity supervision thus updates the shared ViT and query representations without altering the segmentation matching criterion.

\section{\uppercase{Dynamicity Labels}}
\label{sec:offline_labels}

\begin{figure*}[!t]
  \centering
  \resizebox{\textwidth}{!}{
  \begin{tikzpicture}[x=1cm,y=1cm,font=\sffamily,text=black!82]
    \definecolor{plInk}{RGB}{31,37,48}
    \definecolor{plMuted}{RGB}{91,101,116}
    \definecolor{plBlue}{RGB}{43,123,238}
    \definecolor{plBlueFill}{RGB}{236,243,255}
    \definecolor{plGreen}{RGB}{28,157,84}
    \definecolor{plGreenFill}{RGB}{233,248,239}
    \definecolor{plOrange}{RGB}{242,101,34}
    \definecolor{plOrangeFill}{RGB}{255,242,232}
    \definecolor{plRed}{RGB}{211,48,64}
    \definecolor{plGray}{RGB}{124,133,146}
    \definecolor{plGrayFill}{RGB}{245,247,250}

    \tikzset{
      plArrow/.style={
        -{Latex[length=2.0mm,width=1.35mm]},
        draw=plInk,line width=0.78pt
      },
      plBlueArrow/.style={
        -{Latex[length=2.0mm,width=1.35mm]},
        draw=plBlue,line width=0.82pt
      },
      plGreenArrow/.style={
        -{Latex[length=2.0mm,width=1.35mm]},
        draw=plGreen,line width=0.82pt
      },
      plBox/.style={
        rounded corners=1.2pt,line width=0.72pt,
        align=center,font=\scriptsize,text=plInk,
        inner xsep=4pt,inner ysep=3pt
      },
      plImage/.style={
        draw=plInk,line width=0.62pt,inner sep=0pt
      },
      plStep/.style={
        circle,fill=plInk,text=white,
        font=\tiny\bfseries,inner sep=1.6pt
      },
      plTitle/.style={
        font=\scriptsize\bfseries,
        text=plInk,anchor=west
      },
      plChip/.style={
        rounded corners=1.0pt,
        font=\tiny\bfseries,text=white,
        minimum height=0.30cm,inner xsep=3.5pt
      }
    }

    \draw[rounded corners=2pt,draw=plInk!18,
      fill=white,line width=0.55pt]
      (0.15,0.15) rectangle (3.80,3.95);
    \draw[rounded corners=2pt,draw=plInk!18,
      fill=white,line width=0.55pt]
      (4.35,0.15) rectangle (7.35,3.95);
    \draw[rounded corners=2pt,draw=plInk!18,
      fill=white,line width=0.55pt]
      (7.90,0.15) rectangle (11.45,3.95);
    \draw[rounded corners=2pt,draw=plInk!18,
      fill=white,line width=0.55pt]
      (12.00,0.15) rectangle (15.65,3.95);

    \node[plStep] at (0.42,3.68) {1};
    \node[plTitle] at (0.70,3.68)
      {Annotated Video Pair};

    \node[plImage] (frame0) at (1.10,2.45)
      {\includegraphics[
        width=1.60cm,height=0.90cm
      ]{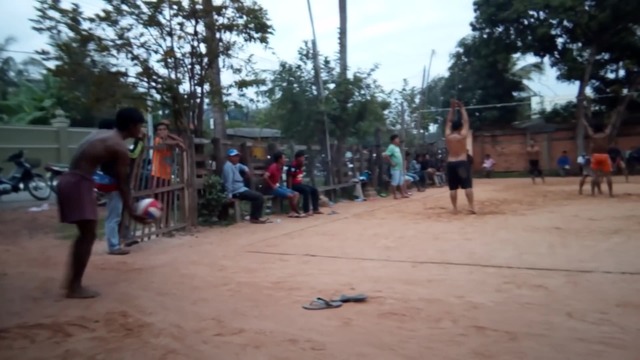}};

    \node[plImage] (frame1) at (2.85,2.45)
      {\includegraphics[
        width=1.60cm,height=0.90cm
      ]{figures/dyneomt_frame_t1.jpg}};

    \draw[plArrow] (frame0.east) -- (frame1.west);

    \node[
      plBox,draw=plGreen,fill=plGreenFill,
      minimum width=2.85cm,minimum height=0.62cm
    ] (mask) at (1.98,1.12)
      {Annotated Mask $M_{j,t}$};

    \node[plStep] at (4.62,3.68) {2};
    \node[plTitle] at (4.90,3.68)
      {Bidirectional Flow};

    \node[plImage] (flow) at (5.85,2.45)
      {\includegraphics[
        width=2.10cm,height=1.18cm
      ]{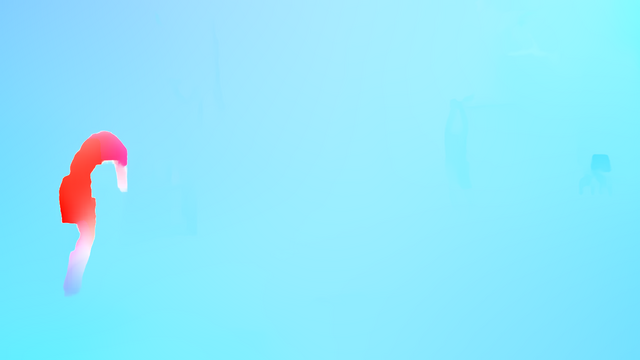}};

    \node[
      plBox,draw=plBlue,fill=plBlueFill,
      minimum width=2.12cm,minimum height=0.62cm
    ] (fb) at (5.85,1.12)
      {$F_t^+,F_t^-$ + FB check};

    \node[plStep] at (8.17,3.68) {3};
    \node[plTitle] at (8.45,3.68)
      {Camera Compensation};

    \node[
      plBox,draw=plOrange,fill=plOrangeFill,
      minimum width=2.60cm,minimum height=0.62cm
    ] (cam) at (9.67,2.68)
      {Robust Camera Field $C_t$};

    \node[font=\scriptsize,text=plMuted]
      at (9.67,2.05)
      {$R_t=F_t^+-C_t$};

    \node[
      plBox,draw=plGreen,fill=plGreenFill,
      minimum width=2.60cm,minimum height=0.62cm
    ] (res) at (9.67,1.12)
      {Residual Flow $R_t$};

    \node[plStep] at (12.27,3.68) {4};
    \node[plTitle] at (12.55,3.68)
      {Motion Evidence and Labels};

    \node[
      plBox,draw=plInk!40,fill=plGrayFill,
      minimum width=2.95cm,minimum height=0.58cm
    ] (score) at (13.83,2.88)
      {Mask-level Score $p_{j,t}$};

    \draw[line width=2.0pt,draw=plGreen]
      (12.55,2.15) -- (13.25,2.15);
    \draw[line width=2.0pt,draw=plGray]
      (13.25,2.15) -- (14.42,2.15);
    \draw[line width=2.0pt,draw=plRed]
      (14.42,2.15) -- (15.12,2.15);

    \node[font=\scriptsize,text=plGreen]
      at (12.85,1.82) {$\leq0.30$};
    \node[font=\scriptsize,text=plGray]
      at (13.83,1.82) {uncertain};
    \node[font=\scriptsize,text=plRed]
      at (14.78,1.82) {$\geq0.70$};

    \node[
      plChip,fill=plGreen,minimum width=0.88cm
    ] at (12.68,1.13) {STATIC};

    \node[
      plChip,fill=plGray,minimum width=0.94cm
    ] at (13.82,1.13) {IGNORE};

    \node[
      plChip,fill=plRed,minimum width=1.05cm
    ] at (15.00,1.13) {DYNAMIC};

    \draw[plBlueArrow]
      (3.80,2.05) -- (4.35,2.05);
    \draw[plBlueArrow]
      (7.35,2.05) -- (7.90,2.05);
    \draw[plGreenArrow]
      (11.45,2.05) -- (12.00,2.05);


  \end{tikzpicture}}

  \caption{Offline dynamicity pseudo-label
  generation. MemFlow estimates bidirectional optical flow between paired frames, and forward--backward consistency filtering removes unreliable correspondences. A robust camera-motion model is fitted to estimate camera-induced flow, which is subtracted from the observed flow. Residual-motion evidence is aggregated within each annotated mask to compute a region-level motion score. The motion score is thresholded to assign dynamic, static, or uncertain targets, while masks with insufficient reliable pixels are directly marked as uncertain.}
  \label{fig:pseudolabels}
\end{figure*}

To supervise the dynamicity head without manually
annotated motion states, we generate class-agnostic
pseudo-labels offline from annotated masks and
paired video frames. As illustrated in Figure~\ref{fig:pseudolabels}, the pipeline combines bidirectional optical-flow estimation and reliability filtering, camera-motion estimation and compensation, mask-level motion scoring, and three-state dynamicity label assignment. For each annotated mask $M_{j,t}$, we assign a label
$y^d_{j,t}\in\{1,0,\varnothing\}$, corresponding to
dynamic, static, or uncertain states, respectively.
The same motion criterion is applied independently of
semantic category, and uncertain targets are ignored
during training.

\subsection{Bidirectional Optical Flow}

We estimate bidirectional optical flow between a reference frame $I_t$ and a paired frame $I_{t'}$. We denote the forward flow from $I_t$ to $I_{t'}$ by $F_t^{+}$ and the backward flow from $I_{t'}$ to $I_t$ by $F_t^{-}$. Let $\Omega$ denote the image domain. For a pixel $x\in\Omega$, its forward-warped location in $I_{t'}$ is

\begin{equation}
x^{+}=x+F_t^{+}(x)
\end{equation}

We sample $F_t^{-}(x^{+})$ by bilinear interpolation. We measure the forward--backward consistency error as

\begin{equation}
e_t^{fb}(x)
=
\left\|
F_t^{+}(x)+F_t^{-}(x^{+})
\right\|_2
\label{eq:fb_error}
\end{equation}

A correspondence is considered reliable only if $x^{+}$ lies within the image domain, both flow vectors are finite, and the consistency error satisfies

\begin{equation}
e_t^{fb}(x)
\leq
1.5+0.05\left\|F_t^{+}(x)\right\|_2
\label{eq:fb_threshold}
\end{equation}

We denote the resulting reliable set by $\Omega_t^{fb}$ and exclude all other pixels from camera-flow estimation and mask-level motion aggregation to limit the influence of occlusions, disocclusions, and unstable flow.

\subsection{Camera-Motion Estimation and Compensation}

Given the reliable correspondences $\Omega_t^{fb}$, we approximate camera-induced image motion using global parametric flow models, without requiring camera intrinsics, depth, or pose. We denote by

$\mathcal{S}_t^{\mathrm{pref}}\subseteq\Omega_t^{fb}$ the preferred support, comprising reliable pixels outside annotated regions. If $\mathcal{S}_t^{\mathrm{pref}}$ contains fewer than 1,000 pixels or covers less than $8\%$ of the image, all reliable correspondences in $\Omega_t^{fb}$ are used instead. We sample at most 24,000 correspondences to form the fitting set $\mathcal{S}_t$.

We fit four candidate flow models $C_t^m$: translation, affine, homography, and quadratic, with $d_m\in\{2,6,8,12\}$ parameters, respectively. The affine and homography models are estimated using RANSAC~\cite{fischler1981ransac}, whereas the quadratic model is fitted by iterative least-squares refitting with MAD-based outlier rejection. For each candidate, the residual is $r_{m,t}(x)=\left\|F_t^{+}(x)-C_t^m(x)\right\|_2$. We select the model by balancing residual error and model complexity

\begin{align}
J_t(m)
={}&
\med\!\left(r_{m,t}\right)
+0.20\,Q_{0.90}\!\left(r_{m,t}\right)
\nonumber\\
&+0.035\,d_m,
\label{eq:modelscore}\\
m_t^*
={}&
\operatorname*{arg\,min}_{m} J_t(m),
\label{eq:modelselection}
\end{align}

where $d_m$ denotes the number of model parameters,
and $\med$ and $Q_{0.90}$ denote the median and
90th percentile computed over the fitting set
$\mathcal{S}_t$. The selected model defines the estimated camera-flow field $C_t(x)=C_t^{m_t^*}(x)$. We subtract this approximation from the observed optical flow to obtain the residual motion field

\begin{equation}
R_t(x)=F_t^{+}(x)-C_t(x).
\label{eq:residual_flow}
\end{equation}

From the residuals of the selected model, we estimate a robust scale and derive an adaptive motion threshold

\begin{align}
\mu_t &= \med r_{m_t^*,t}, \label{eq:mu}\\
\sigma_t &= 1.4826\,\med
\left|r_{m_t^*,t}-\mu_t\right|, \label{eq:sigma}\\
\tau_t &= \max\Big(
\mu_t+3\max(\sigma_t,0.20),\;1.25,
\notag\\
&\qquad\qquad
0.0015\sqrt{H^2+W^2}
\Big).
\label{eq:threshold}
\end{align}

The threshold $\tau_t$ determines significant
camera-compensated residual motion.

\subsection{Mask-Level Motion Evidence}

Each annotated mask $M_{j,t}$ is eroded using an area-adaptive radius of one to six pixels to reduce mask-boundary leakage. We denote by $V_{j,t}$ the set of reliable pixels inside the eroded mask. Over $V_{j,t}$, we compute the median residual magnitude $m_{j,t}$, its 75th percentile $q_{j,t}$, the fraction $a_{j,t}$ of pixels satisfying $\|R_t(x)\|_2>\tau_t$, and the directional coherence

\begin{equation}
k_{j,t}=
\frac{
\left\|
\frac{1}{|V_{j,t}|}
\sum_{x\in V_{j,t}}R_t(x)
\right\|_2
}{
\frac{1}{|V_{j,t}|}
\sum_{x\in V_{j,t}}\|R_t(x)\|_2+\epsilon
}.
\label{eq:directional_coherence}
\end{equation}

We combine these statistics into a motion-evidence score

\begin{gather}
\begin{aligned}
z_{j,t} ={}&
0.55\left(\frac{m_{j,t}}{\tau_t}-1\right)
+0.30\left(\frac{q_{j,t}}{1.35\tau_t}-1\right)
\\
&+0.80(a_{j,t}-0.35)
+0.08\max(k_{j,t}-0.5,0)
\end{aligned}
\label{eq:evidence_z}
\\
p_{j,t}
=
\operatorname{sigmoid}(2.5z_{j,t})
\label{eq:evidence_p}
\end{gather}

The resulting score $p_{j,t}\in[0,1]$ summarizes residual-motion evidence within each annotated mask, with higher values indicating stronger evidence of motion independent of the camera.

\subsection{Three-State Dynamicity Targets}

We assign \emph{dynamic} when $p_{j,t}\geq0.70$ and \emph{static} when $p_{j,t}\leq0.30$. Masks satisfying $0.30<p_{j,t}<0.70$ are marked \emph{uncertain}. This uncertainty interval prevents ambiguous evidence caused by boundary leakage, occlusions, small masks, or optical-flow failures from becoming hard supervision. Masks with fewer than $\max(20,0.03|M_{j,t}|)$ reliable pixels are directly marked as uncertain without computing a motion score. Only dynamic and static targets contribute to the dynamicity loss.

\section{\uppercase{Experimental setup}}
\label{sec:setup}

\subsection{Dynamicity Dataset Construction}

\paragraph{Pseudo-label generation.}
We generate offline dynamicity pseudo-labels using the pipeline described in Section~\ref{sec:offline_labels}. For bidirectional optical-flow estimation, we employ MemFlow~\cite{dong2024memflow} with the publicly available MemFlowNet Sintel checkpoint, applied between consecutive annotated frames. Frames are resized, when necessary, so that the longest side does not exceed 1280 pixels while preserving aspect ratio. The resulting flow fields are then brought back to the annotation resolution by bilinear resizing, with horizontal and vertical components rescaled according to the corresponding spatial factors. The same pseudo-label generation configuration is used across all datasets.

\paragraph{Annotation adaptation.}
We construct dynamicity supervision for the training splits of four video segmentation datasets: VIPSeg~\cite{miao2022vipseg} for video panoptic segmentation, OVIS~\cite{qi2022ovis} and YouTube-VIS 2022~\cite{yang2019ytvis} for video instance segmentation, and VSPW~\cite{miao2021vspw} for video semantic segmentation. The same dynamicity criterion is applied across datasets, with dataset-specific processing only to accommodate differences in annotation structure.  OVIS and YouTube-VIS provide instance tracks directly, whereas VIPSeg panoptic segment identifiers are retained as temporal identities. For VSPW, connected components larger than 64 pixels are extracted within each semantic class and associated across frames by Hungarian matching based on a flow-propagated centroid and an area-change cost. Because the VSPW model predicts one mask per semantic class, valid binary component targets are aggregated into class-level targets using component areas as weights.

\paragraph{Generated supervision.}
Table~\ref{tab:data} summarizes the generated training supervision. An observation corresponds to one annotated mask or connected-component appearance before VSPW class-level aggregation. Across the four training sets, the pipeline produces 3,912,756 valid dynamic or static targets and withholds 394,220 observations as uncertain.

\begin{table}[t]
  \centering
  \caption{Offline dynamicity supervision on training splits. Dynamic rate denotes the proportion of dynamic observations among valid (dynamic or static) labels, excluding uncertain observations.}
  \label{tab:data}
  \begingroup
  \scriptsize
  \setlength{\tabcolsep}{2.0pt}
  \renewcommand{\arraystretch}{0.95}
  \begin{tabular*}{\columnwidth}{@{\extracolsep{\fill}}lrrrrrr}
    \toprule
    Dataset & Videos & Frames & Dynamic & Static & Uncertain & Dyn. rate \\
    \midrule
    \textbf{VIPSeg} & 2,806 & 66,767 & 110,529 & 514,550 & 79,138 & 17.7\% \\
    \cmidrule(lr){1-7}
    \textbf{OVIS} & 607 & 42,149 & 106,619 & 57,608 & 39,186 & 64.9\% \\
    \cmidrule(lr){1-7}
    \textbf{YTVIS22} & 2,985 & 90,160 & 90,633 & 45,789 & 33,437 & 66.4\% \\
    \cmidrule(lr){1-7}
    \textbf{VSPW} & 2,806 & 197,253 & 242,566 & 2,744,462 & 242,459 & 8.1\% \\
    \bottomrule
  \end{tabular*}
  \endgroup
\end{table}

\subsection{Metrics}

We report three metric groups: native video segmentation metrics, conditional dynamicity metrics, and end-to-end dynamic segmentation metrics. Dynamicity evaluation uses automatically generated reference labels on the corresponding evaluation splits.

\paragraph{Original-task preservation.}
To assess the preservation of segmentation performance, we use the standard metrics of each video segmentation task: VPQ~\cite{kim2020vps} and STQ~\cite{weber2021step} for VIPSeg, AP and AR~\cite{yang2019ytvis,qi2022ovis} for OVIS and YouTube-VIS, and mIoU and $\mathrm{mVC}_{16}$~\cite{miao2021vspw} for VSPW.

\paragraph{Conditional dynamicity classification.}
Predicted masks are greedily matched to valid targets by decreasing IoU, requiring the same semantic class and $\mathrm{IoU}\geq0.5$. Dynamicity scores are thresholded at $0.5$, and only matched pairs enter the binary confusion matrix. We report balanced accuracy (BAcc)~\cite{brodersen2010balanced} and MCC~\cite{matthews1975mcc}, which are appropriate for the strongly imbalanced dynamic/static distributions in Table~\ref{tab:data}. These metrics evaluate dynamicity classification conditional on a valid class-and-mask match.

\paragraph{End-to-end dynamic segmentation.}

Because conditional metrics ignore unmatched objects, we also report coverage and DynSeg-F1. DynSeg-F1 counts a success only when mask localization, semantic class, and dynamicity are all correct. Let $G$ denote the number of target masks with valid dynamic or static labels, and $P$ the number of predictions considered for dynamicity evaluation. Let $\mathcal{M}$ denote the class-consistent one-to-one matching set obtained with the same IoU criterion, $M=|\mathcal{M}|$. We denote by $C$ the number of matched pairs with correct dynamicity. Targets labeled as uncertain are excluded from $G$ and from dynamicity matching. In contrast, $P$ includes every non-empty prediction considered by the evaluator, irrespective of its overlap with uncertain targets. Consequently, a prediction that cannot be matched to a valid target, including one located only on an uncertain region, reduces DynSeg precision. We define

\begin{equation}
\begin{aligned}
\mathrm{Coverage} &= \frac{M}{G},\\
\mathrm{DynSeg\!-\!P} &= \frac{C}{P},\qquad
\mathrm{DynSeg\!-\!R} = \frac{C}{G}.
\end{aligned}
\label{eq:coverage_dynseg_pr}
\end{equation}

and

\begin{equation}
\begin{aligned}
\mathrm{DynSeg\!-\!F1}
&=
\frac{
2\,\mathrm{DynSeg\!-\!P}\,\mathrm{DynSeg\!-\!R}
}{
\mathrm{DynSeg\!-\!P}+\mathrm{DynSeg\!-\!R}
}
\\
&=
\frac{2C}{P+G}.
\end{aligned}
\label{eq:dynseg_f1}
\end{equation}

Coverage measures the fraction of valid targets receiving a class-consistent mask match. Unlike conditional BAcc and MCC, which are computed only on matched pairs, DynSeg-F1 provides an end-to-end evaluation: missed valid targets reduce recall, unmatched predictions reduce precision, and matched pairs with incorrect dynamicity are counted as errors. All metrics are reported as percentages, except MCC, which is scaled to $[-100,100]$.

\paragraph{Runtime evaluation.}
Runtime is measured on an NVIDIA Jetson AGX Orin Developer Kit with FP16 inference and batch size one. Following the official VidEoMT benchmark, we compile the backbone with \texttt{torch.compile} and use CUDA events to time only \texttt{model.backbone} after warmup. Image normalization and padding, model post-processing, disk I/O, dataloader work, metric computation, and visualization are outside the timed region. Query propagation, the segmentation heads, and the dynamicity head when present remain inside it.

\subsection{Training Details}

We instantiate DynEoMT with DINOv2 ViT-S/14, ViT-B/14, and ViT-L/14 encoders~ \cite{oquab2024dinov2,norouzi2026videomt}, denoted S, B, and L in the result tables, respectively. All variants use 200 queries, with corresponding query dimensions $D\in\{384,768,1024\}$. ViT-L is initialized from dataset-specific public VidEoMT checkpoints, while ViT-S and ViT-B use the released YouTube-VIS 2019 checkpoints of matching capacity,
with incompatible classification heads randomly initialized. All parameters are trainable.

Training uses three adjacent frames per sample, AdamW~\cite{loshchilov2019adamw}, batch size one, a learning rate of $1.25\times10^{-5}$, weight decay $0.05$, and layer-wise learning-rate decay $0.6$. The dynamicity head uses an $8\times$ learning-rate multiplier. We train for 40k iterations on VIPSeg and VSPW, and 160k on OVIS and YouTube-VIS 2022. Positive-class BCE weights $w_+$ are set to the training static-to-dynamic ratios: $4.655$ for VIPSeg, $0.540$ for OVIS, $0.505$ for YouTube-VIS 2022, and $8.0$ for VSPW, with the latter capped at $8.0$. We use automatic mixed precision and full-model gradient clipping at $0.01$.

For OVIS and YouTube-VIS, track identities are aligned across sampled frames, with $-1$ indicating absence. Each track inherits its class from the first frame where it is visible, avoiding incorrect no-object assignments for tracks entering or leaving the clip. 

For ViT-S and ViT-B, matched VidEoMT baselines are trained with identical initialization, data sampling, optimizer, and training schedule, omitting only the dynamicity head and loss. No auxiliary data are used during target-dataset fine-tuning. For ViT-L, we compare against the published dataset-specific VidEoMT checkpoints used for initialization, as no matched retraining baseline is available.

\section{\uppercase{Results}}
\label{sec:results}

\subsection{Manual Audit of Generated Dynamicity Labels}

To assess the fidelity of the generated supervision, we manually audit a subset of VIPSeg comprising 410 training videos and 6,063 annotated mask tracks. Among these tracks, 2,000 are manually labeled dynamic and 4,063 static, corresponding to a 33.0\% dynamic-track rate. Dynamic tracks account for 11.5\% of the annotated mask area. The offline generator operates at frame level, whereas the manual audit provides a single dynamicity state for each complete track. We therefore evaluate two deterministic track-level aggregations over valid, non-uncertain frame-level pseudo-labels. \emph{Majority} assigns the track the state observed in most valid frames, whereas \emph{any-dynamic} assigns the track as dynamic if at least one valid frame is labeled dynamic. Tracks with no valid frame-level pseudo-label are counted as abstentions.

The generator provides at least one valid frame-level pseudo-label for 5,886 of the 6,063 audited tracks, corresponding to 97.1\% track-level coverage and 2.9\% abstention. Table~\ref{tab:human_audit} shows a precision--recall trade-off between the two aggregation rules. \emph{Majority} yields higher dynamic precision (80.5 vs.\ 66.8) and MCC (57.2 vs.\ 54.3), whereas \emph{any-dynamic} improves recall (72.8 vs.\ 58.9), BAcc (77.7 vs.\ 76.0), and F1 (69.7 vs.\ 68.0).

Overall, the audit indicates substantial agreement with human dynamicity judgments, although performance depends on the temporal aggregation rule. The generated pseudo-labels should
therefore be interpreted as approximate supervision rather than physical-motion ground truth.

\begin{table}[t]
  \centering
  \caption{Human audit of VIPSeg dynamicity pseudo-labels.
  Classification metrics are computed on tracks with valid
  aggregated labels; coverage is measured over all audited
  tracks.}
  \label{tab:human_audit}
  \begingroup
  \scriptsize
  \setlength{\tabcolsep}{2.6pt}
  \renewcommand{\arraystretch}{0.95}
  \begin{tabular*}{\columnwidth}{@{\extracolsep{\fill}}lrrrrrr}
    \toprule
    Aggregation & Coverage & Prec. & Rec. & BAcc & F1 & MCC \\
    \midrule
    Majority & 97.1 & 80.5 & 58.9 & 76.0 & 68.0 & 57.2 \\
    Any-dynamic & 97.1 & 66.8 & 72.8 & 77.7 & 69.7 & 54.3 \\
    \bottomrule
  \end{tabular*}
  \endgroup
\end{table}

\subsection{Preservation of Video Segmentation}

Table~\ref{tab:segmentation} compares DynEoMT with VidEoMT using the native metrics of each segmentation task. For the matched ViT-S/B experiments, introducing dynamicity supervision changes segmentation performance by at most 1.62 points, indicating limited interference with the original task. The ViT-L results differ by at most 1.8 points from the published dataset-specific checkpoints, although this comparison does not isolate the effect of dynamicity training. The lower absolute VIPSeg scores for ViT-S/B reflect their non-target-specific initialization.

\begin{table}[t]
  \centering
  \caption{Native video segmentation performance across datasets and backbone sizes. Values are percentages.}
  \label{tab:segmentation}
  \begingroup
  \scriptsize
  \setlength{\tabcolsep}{2.2pt}
  \renewcommand{\arraystretch}{0.96}
  \begin{tabular*}{\columnwidth}{@{}l@{\extracolsep{\fill}}cccc@{}}
    \toprule
    Dataset & Backbone & Metric & VidEoMT & DynEoMT \\
    \midrule
    \multirow[c]{6}{*}{\textbf{VIPSeg}}
      & \multirow[c]{2}{*}{S} & $\mathrm{VPQ}$ & 4.45 & 4.32 \\
      & & $\mathrm{STQ}$ & 8.24 & 8.05 \\
    \cmidrule(lr){2-5}
      & \multirow[c]{2}{*}{B} & $\mathrm{VPQ}$ & 7.57 & 6.75 \\
      & & $\mathrm{STQ}$ & 12.66 & 11.55 \\
    \cmidrule(lr){2-5}
      & \multirow[c]{2}{*}{L} & $\mathrm{VPQ}$ & 55.7 & 54.5 \\
      & & $\mathrm{STQ}$ & 49.4 & 47.6 \\
    \midrule
    \multirow[c]{6}{*}{\textbf{OVIS}}
      & \multirow[c]{2}{*}{S} & $\mathrm{AP}$ & 26.04 & 25.82 \\
      & & $\mathrm{AR}_{10}$ & 36.57 & 35.80 \\
    \cmidrule(lr){2-5}
      & \multirow[c]{2}{*}{B} & $\mathrm{AP}$ & 39.20 & 37.94 \\
      & & $\mathrm{AR}_{10}$ & 45.92 & 44.30 \\
    \cmidrule(lr){2-5}
      & \multirow[c]{2}{*}{L} & $\mathrm{AP}$ & 51.17 & 50.11 \\
      & & $\mathrm{AR}_{10}$ & 56.55 & 56.19 \\
    \midrule
    \multirow[c]{6}{*}{\textbf{YTVIS22}}
      & \multirow[c]{2}{*}{S} & $\mathrm{AP}$ & 41.40 & 40.09 \\
      & & $\mathrm{AR}_{10}$ & 52.44 & 51.12 \\
    \cmidrule(lr){2-5}
      & \multirow[c]{2}{*}{B} & $\mathrm{AP}$ & 47.55 & 46.51 \\
      & & $\mathrm{AR}_{10}$ & 57.58 & 56.86 \\
    \cmidrule(lr){2-5}
      & \multirow[c]{2}{*}{L} & $\mathrm{AP}$ & 58.86 & 58.64 \\
      & & $\mathrm{AR}_{10}$ & 64.39 & 65.11 \\
    \midrule
    \multirow[c]{6}{*}{\textbf{VSPW}}
      & \multirow[c]{2}{*}{S} & $\mathrm{mVC}_{16}$ & 93.21 & 92.36 \\
      & & $\mathrm{mIoU}$ & 19.74 & 18.46 \\
    \cmidrule(lr){2-5}
      & \multirow[c]{2}{*}{B} & $\mathrm{mVC}_{16}$ & 93.37 & 93.77 \\
      & & $\mathrm{mIoU}$ & 33.29 & 31.91 \\
    \cmidrule(lr){2-5}
      & \multirow[c]{2}{*}{L} & $\mathrm{mVC}_{16}$ & 94.9 & 94.6 \\
      & & $\mathrm{mIoU}$ & 64.8 & 64.6 \\
    \bottomrule
  \end{tabular*}
  \endgroup
\end{table}

\subsection{Dynamicity Prediction}

Table~\ref{tab:dynamic} evaluates dynamicity prediction across datasets and backbone sizes using conditional BAcc/MCC and end-to-end Coverage/DynSeg-F1. ViT-B achieves the highest conditional BAcc on all four datasets, whereas ViT-L consistently obtains the highest Coverage and DynSeg-F1. On VIPSeg, Coverage increases from 25.8\% for ViT-S to 60.7\% for ViT-L, with DynSeg-F1 rising from 29.8\% to 55.0\%. VSPW exhibits a similar trend, highlighting the importance of segmentation coverage for end-to-end dynamicity prediction. However, ViT-L benefits from stronger dataset-specific initialization, so these comparisons do not isolate the effect of backbone capacity. Qualitative predictions on VIPSeg and OVIS are provided in the Appendix.

\begin{table}[t]
  \centering
  \caption{Dynamicity and backbone capacity. BAcc and MCC are conditional on valid mask matches; Coverage and DynSeg-F1 are end-to-end. }
  \label{tab:dynamic}
  \begingroup
  \scriptsize
  \setlength{\tabcolsep}{2.6pt}
  \renewcommand{\arraystretch}{0.95}
  \begin{tabular*}{\columnwidth}{@{}l@{\extracolsep{\fill}}ccccc@{}}
    \toprule
    Dataset & Backbone & BAcc & MCC & Coverage & DynSeg-F1 \\
    \midrule
    \multirow[c]{3}{*}{\textbf{VIPSeg}}
      & S & 83.36 & 59.37 & 25.77 & 29.79 \\
    \cmidrule(lr){2-6}
      & B & 84.30 & 59.46 & 32.98 & 36.48 \\
    \cmidrule(lr){2-6}
      & L & 83.80 & 58.00 & 60.70 & 55.00 \\
    \midrule
    \multirow[c]{3}{*}{\textbf{OVIS}}
      & S & 64.11 & 30.22 & 82.27 & 20.65 \\
    \cmidrule(lr){2-6}
      & B & 67.98 & 36.23 & 85.96 & 22.74 \\
    \cmidrule(lr){2-6}
      & L & 67.60 & 36.59 & 90.48 & 25.22 \\
    \midrule
    \multirow[c]{3}{*}{\textbf{YTVIS22}}
      & S & 66.86 & 32.29 & 86.62 & 15.69 \\
    \cmidrule(lr){2-6}
      & B & 68.59 & 36.93 & 90.97 & 17.43 \\
    \cmidrule(lr){2-6}
      & L & 68.56 & 36.46 & 94.52 & 19.27 \\
    \midrule
    \multirow[c]{3}{*}{\textbf{VSPW}}
      & S & 86.26 & 58.99 & 26.53 & 33.46 \\
    \cmidrule(lr){2-6}
      & B & 87.64 & 58.94 & 32.86 & 39.68 \\
    \cmidrule(lr){2-6}
      & L & 83.40 & 52.70 & 66.50 & 61.10 \\
    \bottomrule
  \end{tabular*}
  \endgroup
\end{table}

\subsection{Computational Overhead}

Table~\ref{tab:runtime} compares the inference throughput of VidEoMT and DynEoMT on the Jetson AGX Orin. The two-layer dynamicity head adds only $256D+513$ parameters (0.099M--0.263M across ViT-S--L). Across the 12 configurations, throughput variations range from a 2.3\% decrease to a 6.1\% increase, with no consistent measured slowdown. These results suggest that the additional dynamicity head introduces little computational overhead within the measured inference path.

\begin{table}[t]
  \centering
  \caption{Backbone runtime on a Jetson AGX Orin across datasets and ViT
  scales. Higher FPS and lower ms/frame are better.}
  \label{tab:runtime}
  \scriptsize
  \setlength{\tabcolsep}{2.3pt}
  \renewcommand{\arraystretch}{0.95}
  \begin{tabular*}{\columnwidth}{@{\extracolsep{\fill}}llcccc}
    \toprule
    Dataset & ViT & \multicolumn{2}{c}{FPS} &
                    \multicolumn{2}{c}{ms/frame} \\
    \cmidrule(lr){3-4}\cmidrule(lr){5-6}
            &     & VidEoMT & DynEoMT & VidEoMT & DynEoMT \\
    \midrule

    \multirow[c]{3}{*}{\textbf{VIPSeg}}
      & S & 28.89 & 28.48 & 34.61 & 35.12 \\
      & B & 13.16 & 13.03 & 75.98 & 76.76 \\
      & L & 4.59 & 4.74 & 217.72 & 210.87 \\
    \cmidrule(lr){1-6}

    \multirow[c]{3}{*}{\textbf{OVIS}}
      & S & 42.70 & 41.89 & 23.42 & 23.87 \\
      & B & 21.64 & 21.65 & 46.21 & 46.18 \\
      & L & 8.32 & 8.35 & 120.21 & 119.69 \\
    \cmidrule(lr){1-6}

    \multirow[c]{3}{*}{\textbf{YTVIS22}}
      & S & 57.14 & 60.64 & 17.50 & 16.49 \\
      & B & 33.33 & 33.74 & 30.01 & 29.64 \\
      & L & 13.01 & 13.01 & 76.89 & 76.85 \\
    \cmidrule(lr){1-6}

    \multirow[c]{3}{*}{\textbf{VSPW}}
      & S & 26.46 & 25.86 & 37.79 & 38.67 \\
      & B & 13.07 & 13.00 & 76.53 & 76.95 \\
      & L & 4.61 & 4.73 & 217.01 & 211.53 \\
    \bottomrule
  \end{tabular*}
\end{table}

\subsection{Probing Dynamicity in Segmentation Queries}

To investigate whether VidEoMT queries encode dynamicity information, we evaluate linear and MLP probes on frozen ViT-L representations from VIPSeg. We also assess the effect of query propagation and compare against the fully fine-tuned DynEoMT model.

\begin{table}[t]
\centering
\small
\caption{Dynamicity probing on VIPSeg using ViT-L.
Comparison of frozen query representations,
query propagation, and joint fine-tuning.}
\label{tab:query_probe}
\begingroup
\scriptsize
\setlength{\tabcolsep}{3.0pt}
\begin{tabular*}{\columnwidth}{@{\extracolsep{\fill}}lrrrr}
\toprule
Setting & BAcc & MCC & Cov. & DynSeg-F1 \\
\midrule
Frozen + linear probe & 81.56 & 50.48 & 61.91 & 51.53 \\
Frozen + MLP probe & 83.75 & 55.46 & 61.91 & 53.83 \\
Frozen + MLP, no propagation & 84.82 & 57.67 & 54.47 & 51.92 \\
Full DynEoMT & 83.80 & 58.00 & 60.70 & 55.00 \\
\bottomrule
\end{tabular*}
\endgroup
\end{table}

Table~\ref{tab:query_probe} shows that frozen segmentation queries already support dynamicity prediction. The MLP probe improves BAcc from 81.56 to 83.75 and MCC from 50.48 to 55.46 compared with the linear probe. Its conditional BAcc is also close to that of fully fine-tuned DynEoMT (83.75 vs. 83.80). Removing query propagation improves conditional BAcc to 84.82 but reduces Coverage from 61.91 to 54.47 and DynSeg-F1 from 53.83 to 51.92. Thus, propagation improves end-to-end performance without improving conditional classification. Full fine-tuning achieves the highest MCC and DynSeg-F1, indicating a benefit from jointly optimizing the shared representation and the dynamicity head.

\subsection{Ablation studies}

\subsubsection{Ablation of Camera-Motion Compensation}

We assess the contribution of camera-motion compensation to pseudo-label quality on the manually audited VIPSeg subset, using the generation pipeline described in Section~\ref{sec:offline_labels} as reference. We compare the full pseudo-label generation pipeline with a variant that omits camera-flow compensation. Both variants are evaluated using majority and any-dynamic track-level aggregation.

\begin{table}[t]
\centering
\caption{Impact of camera-motion compensation on
pseudo-label quality on the manually audited VIPSeg
subset. Full denotes the complete pipeline; No
comp. removes camera-motion compensation.}
\label{tab:camera_ablation}
\begingroup
\scriptsize
\setlength{\tabcolsep}{3pt}
\renewcommand{\arraystretch}{1.05}
\begin{tabular*}{\columnwidth}{
@{\extracolsep{\fill}}lrrrr@{}}
\toprule
& \multicolumn{2}{c}{Majority}
& \multicolumn{2}{c}{Any-dynamic} \\
\cmidrule(lr){2-3}
\cmidrule(lr){4-5}
Metric & Full & No comp.
       & Full & No comp. \\
\midrule
Coverage & 97.1 & 98.6 & 97.1 & 98.6 \\
Precision & 80.5 & 90.7 & 66.8 & 81.2 \\
Recall    & 58.9 & 28.9 & 72.8 & 47.8 \\
BAcc      & 76.0 & 63.7 & 77.7 & 71.2 \\
F1        & 68.0 & 43.8 & 69.7 & 60.2 \\
MCC       & 57.2 & 42.2 & 54.3 & 50.5 \\
\bottomrule
\end{tabular*}
\endgroup
\end{table}

Table~\ref{tab:camera_ablation} shows that camera-motion compensation substantially improves the detection of independently moving regions. Without compensation, majority-aggregation BAcc decreases from 76.0 to 63.7, while dynamic-class F1 drops from 68.0 to 43.8. Dynamic recall falls from 58.9 to 28.9, despite an increase in precision. A similar trend is observed with any-dynamic aggregation, where BAcc decreases from 77.7 to 71.2 and F1 from 69.7 to 60.2.

These results highlight the importance of separating camera-induced image motion from residual object motion when constructing dynamicity supervision. Without compensation, the generator becomes more conservative in identifying moving regions, substantially reducing dynamic recall.

\subsubsection{Ablation of the Uncertainty Interval}

We investigate the effect of the uncertainty interval used to assign dynamic and static pseudo-labels. The default thresholds $(0.30, 0.70)$ are compared with a narrower interval $(0.35, 0.65)$ and a wider interval $(0.25, 0.75)$, while keeping all other parameters unchanged. Evaluation is performed on the manually audited VIPSeg subset using majority track-level aggregation.

\begin{table}[t]
\centering
\caption{Effect of the uncertainty interval on
pseudo-label quality. Results use majority track-level
aggregation on the audited VIPSeg subset. All values
are percentages, except MCC, which is scaled by 100.}
\label{tab:uncertainty_ablation}
\begingroup
\scriptsize
\setlength{\tabcolsep}{3pt}
\renewcommand{\arraystretch}{1.05}
\begin{tabular*}{\columnwidth}{
@{\extracolsep{\fill}}lrrrr@{}}
\toprule
Thresholds & Coverage & BAcc & F1 & MCC \\
\midrule
$(0.35, 0.65)$ & 97.6 & 75.8 & 67.7 & 57.0 \\
$(0.30, 0.70)$ & 97.1 & 76.0 & 68.0 & 57.2 \\
$(0.25, 0.75)$ & 96.0 & 76.4 & 68.5 & 57.9 \\
\bottomrule
\end{tabular*}
\endgroup
\end{table}

Table~\ref{tab:uncertainty_ablation} illustrates the trade-off between pseudo-label coverage and classification reliability. Widening the uncertainty interval improves BAcc from 76.0 to 76.4 and F1 from 68.0 to 68.5, but reduces coverage from 97.1\% to 96.0\%. Conversely, narrowing the interval slightly increases coverage while reducing classification performance. The differences remain modest across the tested configurations, suggesting that pseudo-label quality is not highly sensitive to these thresholds. The default interval provides a compromise between retaining supervision and excluding ambiguous motion evidence.

\subsubsection{Effect of Forward--Backward Consistency Filtering}

We evaluate the contribution of forward--backward consistency filtering to pseudo-label quality on the manually audited VIPSeg subset. We compare the complete pseudo-label generation pipeline with a variant that disables this filtering step, while keeping the remaining components unchanged.

\begin{table}[t]
\centering
\caption{Effect of forward--backward consistency
filtering on VIPSeg pseudo-label quality, using
majority track-level aggregation.}
\label{tab:fb_ablation}
\begingroup
\scriptsize
\setlength{\tabcolsep}{3pt}
\renewcommand{\arraystretch}{1.05}
\begin{tabular*}{\columnwidth}{
@{\extracolsep{\fill}}lrrrr@{}}
\toprule
Configuration & Cov. & BAcc & F1 & MCC \\
\midrule
Full pipeline & 97.1 & 76.0 & 68.0 & 57.2 \\
Without FB filtering & 99.2 & 75.5 & 67.3 & 56.2 \\
\bottomrule
\end{tabular*}
\endgroup
\end{table}

As shown in Table~\ref{tab:fb_ablation}, removing forward--backward consistency filtering increases coverage from 97.1\% to 99.2\%, but slightly reduces BAcc from 76.0 to 75.5 and F1 from 68.0 to 67.3. MCC also decreases from 57.2 to 56.2. These results indicate that consistency filtering provides a modest improvement in pseudo-label reliability at the expense of coverage. Its effect is substantially smaller than that of camera-motion compensation.


\section{Limitations and Discussion}
\label{sec:limitations}

\paragraph{Pseudo-label supervision.}
Camera-compensated optical flow provides approximate motion supervision rather than physical-motion ground truth. Flow estimation errors, occlusions, and reflections can affect label reliability, while axial motion may produce little image displacement and parallax can generate residual motion in static regions. The human audit quantifies agreement with manual judgments but does not eliminate these limitations.

\paragraph{Online dynamicity prediction.}
DynEoMT relies exclusively on the current frame and propagated queries, without explicit motion or geometric inputs at inference. Although these representations contain information predictive of dynamicity, they do not provide an explicit geometric decomposition of camera and object motion. Furthermore, end-to-end dynamicity performance remains dependent on the quality of the underlying segmentation.

\paragraph{Evaluation and generalization.}
In semantic segmentation, disconnected same-class regions with different motion states are aggregated into a single class-level target, as in VSPW. Moreover, pseudo-label generation uses fixed thresholds across datasets, while dynamicity predictions are binarized at $0.5$. Calibration on human-verified validation data could improve robustness across different scenes and motion conditions.

\section{\uppercase{Conclusions}}
\label{sec:conclusion}

We introduced DynEoMT, an online framework that extends query-based video segmentation with region-level dynamicity prediction. Trained using class-agnostic pseudo-labels derived from camera-compensated optical flow, DynEoMT predicts dynamicity directly from propagated segmentation queries, without explicit motion or geometric inputs at inference. Experiments across semantic, instance, and panoptic video segmentation demonstrate that this additional attribute can be learned while largely preserving segmentation performance and introducing minimal overhead in the measured inference path. Our probing experiments further show that dynamicity-relevant information is already present in segmentation queries, while end-to-end performance remains closely tied to segmentation quality.




\begin{thebibliography}{}

\bibitem[Bescos et~al., 2018]{bescos2018dynaslam}
Bescos, B., F\'acil, J.~M., Civera, J., and Neira, J. (2018).
\newblock {DynaSLAM}: Tracking, mapping, and inpainting in dynamic scenes.
\newblock {\em IEEE Robotics and Automation Letters}, 3(4):4076--4083.

\bibitem[Brodersen et~al., 2010]{brodersen2010balanced}
Brodersen, K.~H., Ong, C.~S., Stephan, K.~E., and Buhmann, J.~M. (2010).
\newblock The balanced accuracy and its posterior distribution.
\newblock In {\em ICPR}.

\bibitem[Carion et~al., 2020]{carion2020detr}
Carion, N., Massa, F., Synnaeve, G., Usunier, N., Kirillov, A., and Zagoruyko,
  S. (2020).
\newblock End-to-end object detection with transformers.
\newblock In {\em ECCV}.

\bibitem[Choudhuri et~al., 2023]{choudhuri2023caroq}
Choudhuri, A., Chowdhary, G., and Schwing, A.~G. (2023).
\newblock Context-aware relative object queries to unify video instance and
  panoptic segmentation.
\newblock In {\em Proceedings of the IEEE/CVF Conference on Computer Vision and
  Pattern Recognition}, pages 6377--6386.

\bibitem[Dong and Fu, 2024]{dong2024memflow}
Dong, Q. and Fu, Y. (2024).
\newblock {MemFlow}: Optical flow estimation and prediction with memory.
\newblock In {\em CVPR}.

\bibitem[Fischler and Bolles, 1981]{fischler1981ransac}
Fischler, M.~A. and Bolles, R.~C. (1981).
\newblock Random sample consensus: A paradigm for model fitting with
  applications to image analysis and automated cartography.
\newblock {\em Communications of the ACM}, 24(6):381--395.

\bibitem[Galagain et~al., 2026]{GALAGAIN2026105715}
Galagain, C., Poreba, M., and Goulette, F. (2026).
\newblock Is semantic slam ready for embedded systems? a comparative survey.
\newblock {\em Robotics and Autonomous Systems}, 206:105715.

\bibitem[Goli et~al., 2025]{goli2025romo}
Goli, L., Sabour, S., Matthews, M.~J., Brubaker, M.~A., Lagun, D., Jacobson,
  A., Fleet, D.~J., Saxena, S., and Tagliasacchi, A. (2025).
\newblock {RoMo}: Robust motion segmentation improves structure from motion.
\newblock In {\em Proceedings of the IEEE/CVF International Conference on
  Computer Vision}, pages 6155--6164.

\bibitem[He et~al., 2022]{he2022inspro}
He, F., Zhang, H., Gao, N., Jia, J., Shan, Y., Zhao, X., and Huang, K. (2022).
\newblock {InsPro}: Propagating instance query and proposal for online video
  instance segmentation.
\newblock In {\em Advances in Neural Information Processing Systems},
  volume~35, pages 19370--19383.

\bibitem[He et~al., 2026]{he2026geomotion}
He, X., Lin, P., Cui, Y., Guo, D., Shen, C., and Zhang, X. (2026).
\newblock {GeoMotion}: Rethinking motion segmentation via latent {4D} geometry.
\newblock In {\em Proceedings of the IEEE/CVF Conference on Computer Vision and
  Pattern Recognition}.

\bibitem[Huang et~al., 2022]{huang2022minvis}
Huang, D.-A., Yu, Z., and Anandkumar, A. (2022).
\newblock {MinVIS}: A minimal video instance segmentation framework without
  video-based training.
\newblock In {\em NeurIPS}.

\bibitem[Huang et~al., 2025]{huang2025segmentanymotion}
Huang, N., Zheng, W., Xu, C., Keutzer, K., Zhang, S., Kanazawa, A., and Wang,
  Q. (2025).
\newblock Segment any motion in videos.
\newblock In {\em Proceedings of the IEEE/CVF Conference on Computer Vision and
  Pattern Recognition}, pages 3406--3416.

\bibitem[Jiao et~al., 2021]{jiao2021effiscene}
Jiao, Y., Tran, T.~D., and Shi, G. (2021).
\newblock {EffiScene}: Efficient per-pixel rigidity inference for unsupervised
  joint learning of optical flow, depth, camera pose and motion segmentation.
\newblock In {\em Proceedings of the IEEE/CVF Conference on Computer Vision and
  Pattern Recognition}, pages 5538--5547.

\bibitem[Karazija et~al., 2024]{karazija2024learning}
Karazija, L., Laina, I., Rupprecht, C., and Vedaldi, A. (2024).
\newblock Learning segmentation from point trajectories.
\newblock In {\em Advances in Neural Information Processing Systems},
  volume~37, pages 112573--112597.

\bibitem[Kerssies et~al., 2025]{kerssies2025eomt}
Kerssies, T., Cavagnero, N., Hermans, A., Norouzi, N., Averta, G., Leibe, B.,
  Dubbelman, G., and de~Geus, D. (2025).
\newblock Your {ViT} is secretly an image segmentation model.
\newblock In {\em CVPR}.

\bibitem[Kim et~al., 2020]{kim2020vps}
Kim, D., Woo, S., Lee, J.-Y., and Kweon, I.~S. (2020).
\newblock Video panoptic segmentation.
\newblock In {\em CVPR}.

\bibitem[Kirillov et~al., 2023]{kirillov2023segmentanything}
Kirillov, A., Mintun, E., Ravi, N., Mao, H., Rolland, C., Gustafson, L., Xiao,
  T., Whitehead, S., Berg, A.~C., Lo, W.-Y., Doll{\'a}r, P., and Girshick, R.
  (2023).
\newblock Segment anything.
\newblock In {\em Proceedings of the IEEE/CVF International Conference on
  Computer Vision}, pages 4015--4026.

\bibitem[Lee et~al., 2025]{lee2025cavis}
Lee, S., Seo, J., Han, K., Choi, M., and Im, S. (2025).
\newblock Context-aware video instance segmentation.
\newblock In {\em ICCV}.

\bibitem[Loshchilov and Hutter, 2019]{loshchilov2019adamw}
Loshchilov, I. and Hutter, F. (2019).
\newblock Decoupled weight decay regularization.
\newblock In {\em ICLR}.

\bibitem[Matthews, 1975]{matthews1975mcc}
Matthews, B.~W. (1975).
\newblock Comparison of the predicted and observed secondary structure of {T4}
  phage lysozyme.
\newblock {\em Biochimica et Biophysica Acta}, 405(2):442--451.

\bibitem[Miao et~al., 2022]{miao2022vipseg}
Miao, J., Wang, X., Wu, Y., Li, W., Zhang, X., Wei, Y., and Yang, Y. (2022).
\newblock Large-scale video panoptic segmentation in the wild: A benchmark.
\newblock In {\em CVPR}.

\bibitem[Miao et~al., 2021]{miao2021vspw}
Miao, J., Wei, Y., Wu, Y., Liang, C., Li, G., and Yang, Y. (2021).
\newblock {VSPW}: A large-scale dataset for video scene parsing in the wild.
\newblock In {\em CVPR}.

\bibitem[Norouzi et~al., 2026]{norouzi2026videomt}
Norouzi, N., Zulfikar, I.~E., Cavagnero, N., Kerssies, T., Leibe, B.,
  Dubbelman, G., and de~Geus, D. (2026).
\newblock {VidEoMT}: Your {ViT} is secretly also a video segmentation model.
\newblock In {\em CVPR}.

\bibitem[Oquab et~al., 2024]{oquab2024dinov2}
Oquab, M., Darcet, T., Moutakanni, T., Vo, H., Szafraniec, M., Khalidov, V.,
  Fernandez, P., Haziza, D., Massa, F., El-Nouby, A., et~al. (2024).
\newblock {DINOv2}: Learning robust visual features without supervision.
\newblock {\em TMLR}.

\bibitem[Qi et~al., 2022]{qi2022ovis}
Qi, J., Gao, Y., Hu, Y., Wang, X., Liu, X., Bai, X., Belongie, S., Yuille, A.,
  Torr, P. H.~S., and Bai, S. (2022).
\newblock Occluded video instance segmentation: A benchmark.
\newblock {\em IJCV}, 130(8):2022--2039.

\bibitem[Ravi et~al., 2025]{ravi2025sam2}
Ravi, N., Gabeur, V., Hu, Y.-T., Hu, R., Ryali, C., Ma, T., Khedr, H.,
  R{\"a}dle, R., Rolland, C., Gustafson, L., Mintun, E., Pan, J., Alwala,
  K.~V., Carion, N., Wu, C.-Y., Girshick, R., Doll{\'a}r, P., and
  Feichtenhofer, C. (2025).
\newblock {SAM 2}: Segment anything in images and videos.
\newblock In {\em Proceedings of the International Conference on Learning
  Representations}.

\bibitem[Teed and Deng, 2020]{teed2020raft}
Teed, Z. and Deng, J. (2020).
\newblock {RAFT}: Recurrent all-pairs field transforms for optical flow.
\newblock In {\em ECCV}.

\bibitem[Weber et~al., 2021]{weber2021step}
Weber, M., Xie, J., Collins, M., Zhu, Y., Voigtlaender, P., Adam, H., Green,
  B., Geiger, A., Leibe, B., Cremers, D., Osep, A., Leal-Taix{\'e}, L., and
  Chen, L.-C. (2021).
\newblock {STEP}: Segmenting and tracking every pixel.
\newblock In {\em Proceedings of the Neural Information Processing Systems
  Track on Datasets and Benchmarks}, volume~1.

\bibitem[Xie et~al., 2026]{xie2026gmos}
Xie, J., Han, T., Xie, W., and Zisserman, A. (2026).
\newblock {GMOS}: Grounding moving object segmentation in {3D} space and time.
\newblock {\em arXiv preprint arXiv:2605.30352}.

\bibitem[Xie et~al., 2024]{xie2024flowsam}
Xie, J., Yang, C., Xie, W., and Zisserman, A. (2024).
\newblock Moving object segmentation: All you need is {SAM} (and flow).
\newblock In {\em Proceedings of the Asian Conference on Computer Vision},
  pages 291--308.

\bibitem[Yang et~al., 2021]{yang2021motiongrouping}
Yang, C., Lamdouar, H., Lu, E., Zisserman, A., and Xie, W. (2021).
\newblock Self-supervised video object segmentation by motion grouping.
\newblock In {\em Proceedings of the IEEE/CVF International Conference on
  Computer Vision}, pages 7157--7168.

\bibitem[Yang and Ramanan, 2021]{yang2021rigidmask}
Yang, G. and Ramanan, D. (2021).
\newblock Learning to segment rigid motions from two frames.
\newblock In {\em Proceedings of the IEEE/CVF Conference on Computer Vision and
  Pattern Recognition}, pages 1266--1275.

\bibitem[Yang et~al., 2019]{yang2019ytvis}
Yang, L., Fan, Y., and Xu, N. (2019).
\newblock Video instance segmentation.
\newblock In {\em ICCV}.

\bibitem[Zhang et~al., 2023]{zhang2023dvis}
Zhang, T., Tian, X., Wu, Y., Ji, S., Wang, X., Zhang, Y., and Wan, P. (2023).
\newblock {DVIS}: Decoupled video instance segmentation framework.
\newblock In {\em ICCV}.

\bibitem[Zhang et~al., 2025]{zhang2025dvispp}
Zhang, T., Tian, X., Zhou, Y., Ji, S., Wang, X., Tao, X., Zhang, Y., Wan, P.,
  Wang, Z., and Wu, Y. (2025).
\newblock {DVIS++}: Improved decoupled framework for universal video
  segmentation.
\newblock {\em IEEE TPAMI}.

\bibitem[Zhou et~al., 2024]{zhou2024dvisdaq}
Zhou, Y., Zhang, T., Ji, S., Yan, S., and Li, X. (2024).
\newblock Improving video segmentation via dynamic anchor queries.
\newblock In {\em ECCV}.

\end{thebibliography}


\section*{\uppercase{Appendix}}

\noindent\textbf{Qualitative Results.}
The examples below illustrate DynEoMT predictions on OVIS and VIPSeg. From left to right, the columns alternate between segmentation predictions and corresponding dynamicity maps. Green indicates static regions, while red indicates dynamic regions.

\begin{figure*}[!t]
    \centering
    \includegraphics[width=\textwidth]{
        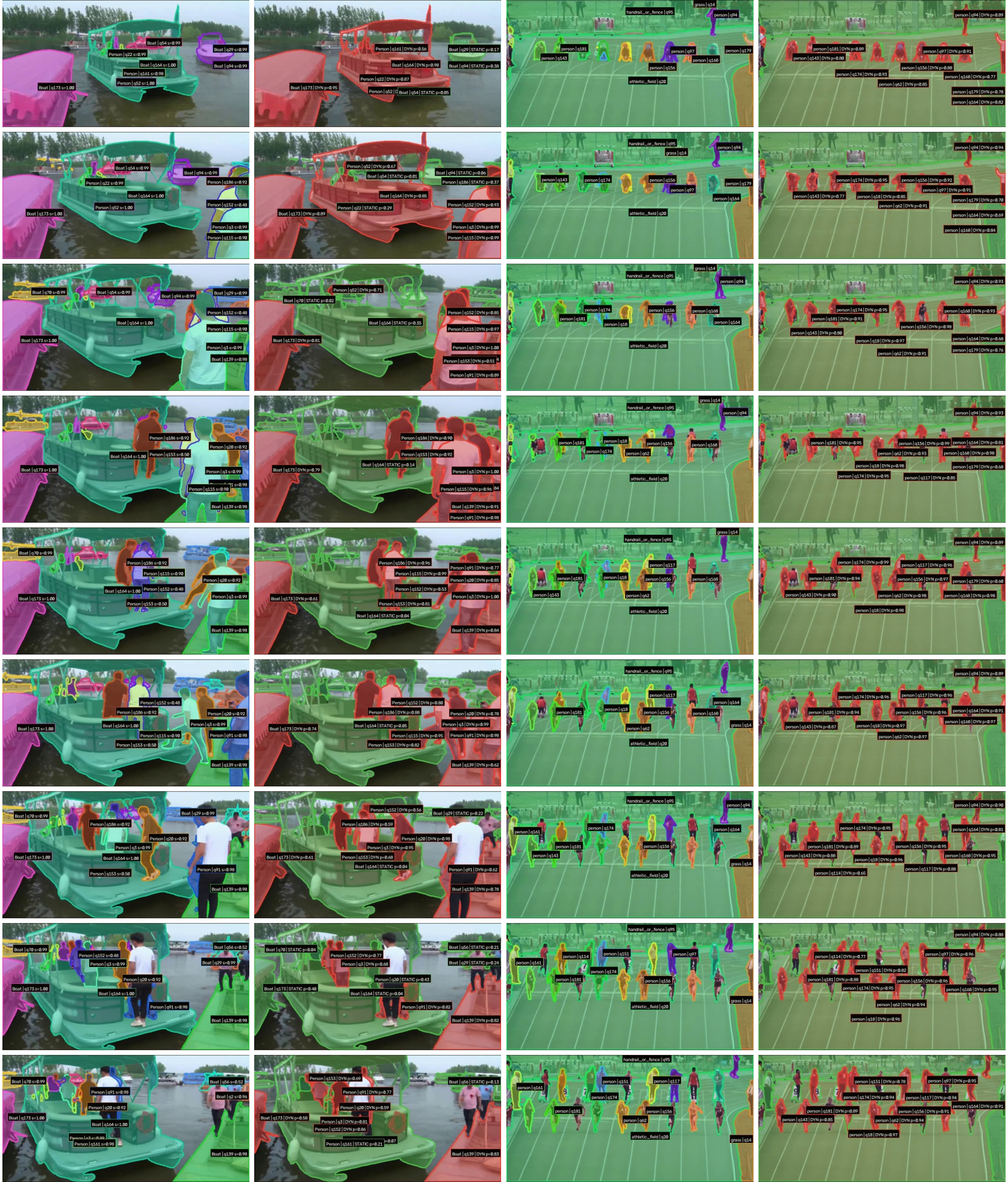}
\end{figure*}

\end{document}